\documentclass[11pt]{article}

\usepackage{dialogue}

\usepackage{ifxetex}
\ifxetex
    \usepackage{fontspec}
    \setromanfont{Times New Roman}
\else
  \usepackage{cmap}
  \usepackage[T2A,T1]{fontenc}
  \usepackage[utf8]{inputenc}
  \usepackage{times}
  \usepackage{latexsym}
  \DeclareFontFamilySubstitution{T2A}{\familydefault}{cmr}
\fi

\usepackage[russian,british]{babel}
\usepackage{url}
\usepackage{pgf}

\usepackage{amssymb}
\usepackage{amsmath}
\usepackage{multirow}
\usepackage{multicol}
\usepackage{subcaption}

\usepackage{covington}

\usepackage{hyperref}

\dialogfinalcopy

\title{Methods for Recognizing Nested Terms}

\author{Igor Rozhkov \\
  Lomonosov Moscow State University \\
  Leninskie Gory, 1/4, Moscow,  Russia \\
  {\tt fulstocky@gmail.com} \\\And
  Natalia Loukachevitch \\
  Lomonosov Moscow State University \\
  Leninskie Gory, 1/4, Moscow,  Russia \\
  {\tt louk\_nat@mail.ru} \\}

\date{}

\begin{document}
\maketitle

\begin{abstract}
In this paper, we describe our participation in the RuTermEval competition devoted to extracting nested terms. We apply the Binder model, which was previously successfully applied to the recognition of nested named entities, to extract nested terms. We obtained the best results of term recognition in all three tracks of the RuTermEval competition. In addition, we study the new task of recognition of nested terms from flat training data annotated with terms without nestedness. We can conclude that several approaches we proposed in this work are viable enough to retrieve nested terms effectively without nested labeling of them.


\end{abstract}

\selectlanguage{russian}
\begin{center}
  \russiantitle{Методы распознавания вложенных терминов}

  \medskip \setlength\tabcolsep{0.5cm}
  \begin{tabular}{cc}
    \textbf{Рожков И.С.} & \textbf{Лукашевич Н.В.}\\
      МГУ имени М.В. Ломоносова&МГУ имени М.В. Ломоносова\\
      Ленинские горы, 1/4, Москва, Россия& Ленинские горы, 1/4, Москва, Россия \\
      {\tt fulstocky@gmail.com} &  {\tt louk\_nat@mail.ru}
  \end{tabular}
  \medskip
\end{center}

\begin{abstract}
 В этой статье мы описываем наше участие в конкурсе RuTermEval, посвященном извлечению вложенных терминов. Мы применяли модель Binder, которая ранее успешно применялась для распознавания вложенных именованных сущностей, для извлечения вложенных терминов. Мы получили наилучшие результаты распознавания терминов во всех трех треках конкурса RuTermEval. Кроме того, мы исследуем новую задачу распознавания вложенных терминов из плоских обучающих данных, аннотированных терминами без вложенности. Несколько методов, предложенных нами в данной работе, показывают, что вложенные термины возможно извлекать эффективно без наличия их вложенной разметки. 
 
  \textbf{Ключевые слова:} вложенные термины, автоматическое извлечение терминов, контрастивное обучение 
\end{abstract}
\selectlanguage{british}

\selectlanguage{british}

\section{Introduction}
\label{intro}

Automatic term extraction is a well-known research task that has been studied for decades.
 Terms are defined as words or phrases that denote concepts of a specific domain, and knowing them is important for domain analysis, machine translation, or domain-specific information retrieval. 
Various approaches have been proposed for automatic term extraction.  However,  automatic methods do not yet achieve the quality of manual term analysis.

During recent years, machine learning methods have been intensively studied \cite{loukachevitch2012automatic,charalampakisComparisonSemisupervisedSupervised2016,nadifUnsupervisedSelfsupervisedDeep2021}.  The application of machine learning improves the quality of term extraction, but requires creating training datasets. In addition, the transfer of a trained model from one domain to another usually leads to degradation of the performance of term extraction. Currently, language models \cite{xiePretrainedLanguageModels2022,liuFinBERTPretrainedFinancial2020}  are texted in automatic term extraction.

Available datasets usually contain so-called flat term annotation, that is, an annotated term cannot include other terms. Another formulation of the task
assumes that terms can be contained within other terms, such terms are called nested terms. For example,  the term "‘infection of upper tract" includes the following term: "infection", "tract", "upper tract". The extraction of nested terms allows the researcher to recognize more terms, but requires the application of special methods.

In 2024 new Russian datasets have been prepared for automatic term analysis in the framework of the RuTerm-2024 evaluation. The datasets are annotated with nested terms. The dataset includes texts from different domains, which allows for the study of model transfer between domains.

In this paper, we consider an approach for the extraction of nested terms and test it in the RuTerm-2024 dataset. In addition, we study the task of the transfer from flat terms to nested ones, that is, we suggest that the training data are annotated only with flat terms, but the trained model should can annotated nested terms. For both tasks, we experiment with the Binder model, which creates representations for target entity types using contrastive learning \footnote{Code is available at \texttt{\url{https://github.com/fulstock/Methods-for-Recognizing-Nested-Terms}}} \cite{zhangOptimizingBiEncoderNamed2023}.

\section{Related Work}
\label{related}

The automatic term extraction task has two main variants: corpus-oriented and document-oriented \cite{vsajatovic2019evaluating}. In the former case, a system should generate a list of terms for a document corpus; in the latter case, the task is to identify mentions of terms in domain documents. This variant can be classified as a sequence labeling task. 

The first corpus-oriented methods utilized statistical measures such as tf-idf, context measures (c-vaule), association measures based on frequencies of component words and co-occurrence frequencies of a candidate phrase \cite{astrakhantsev2015methods}. The combination of measures leads to a significant improvement in term extraction performance \cite {bolshakova2013topic,loukachevitch2012automatic}, but the transfer of a model trained for one domain to another domain leads to degradation of results \cite{hazem2022cross,loukachevitch2013experimental,bolshakovaexperimental}.

In document-oriented term extraction, the authors of \cite{bolshakova2015heuristic} exploited lexico-syntactic patterns and rules. Currently, the most effective approaches to document-oriented term extraction utilized BERT-based techniques \cite{lang2021transforming,tran2024can}.

The extraction of nested terms was studied in several works \cite{marciniak2015nested,tran2024can,vo2022automatic}. The authors of \cite{marciniak2015nested} divide longer terms into syntactically correct component phrases and estimate the "unithood" of candidate terms using the NPMI statistical measure (Normalized Pointwise Mutual Information) \cite{bouma2009normalized}.  Tran et al. \cite{tran2024can}  propose using a novel term annotation scheme for training models. The proposed scheme comprises an additional encoding for nested single terms. Applying the scheme and  XLMR classifier,  the authors obtain the best results on the ACTER multilingual dataset  \cite{rigouts2022acter}.

In a similar task of recognition of nested named entities, various approaches have been proposed \cite{loukachevitch2024nerel,zhangOptimizingBiEncoderNamed2023,zhuRecognizingNestedEntities2022}. One of the best methods for nested named entity recognition is Binder \cite{zhangOptimizingBiEncoderNamed2023}, based on contrastive learning.

\section{RuTermEval Competition and Dataset}

As part of the Dialogue Evaluation in 2025, the RuTermEval competition was organized. In this competition, one should design a model capable of extracting nested terms given the labeled data. 

There are three tracks in this competition:
\begin{itemize}
    \item Track 1: Identification of terms
    \item Track 2: Identification and classification of terms into three classes (specific\_term, common\_term, nomen);
    \item Track 3: Identification and classification of terms into three classes (specific\_term, common\_term, nomen) with the formulation of the transfer learning task to other domains.
\end{itemize}

The organizers used Codalab to hold the competition. In the first track they used the usual F1 score for measuring the teams' submissions accuracy. For the second and third tracks, organizers used weighted F1 and class-agnostic F1 scores. The weighted F1 considers some classes more important than the others, thus increasing their value ratio in the total score. The second one treats the task of term extracting as term identification task, i.e. measuring the accuracy of term prediction regardless of their class. 

The total table of competitors' results was sorted based on the F1 score in the first track and the weighted F1 score in the second and third tracks. 

Three types of terms were considered:
\begin{itemize}
    \item \textit{specific}: terms that are specific both in-domain and lexically;
    \item \textit{common}: terms that are specific only in-domain (they can be known and used by non-specialists);
    \item \textit{nomen}: names of unique objects that belong to a specific domain.
\end{itemize}

Examples for each of three term types are as follows: 
\begin{examples}
\selectlanguage{russian}
\item 
    \gll эпистемической модальности
         epistemic modality
    \glt `Example of specific term'
    \glend
\item 
    \gll пользователю
         user
    \glt `Example of common term'
    \glend
\item 
    \gll Национального корпуса ~ русского языка
         National Corpus of Russian Language
    \glt `Example of nomen term'
    \glend
\selectlanguage{british}
\end{examples}

\begin{table}[t]
\begin{center}
\begin{tabular}{|c|c|c|c|}
\hline
& train & dev (track 1 \& 2) & dev (track 3) \\ \hline
specific & 12664 (69.95\%) & 3387 (67.77\%) & 3270 (58.52\%) \\ 
common & 4866 (26.88\%) & 1275 (25.51\%) & 1173 (20.49\%)\\ 
nomen & 573 (3.17\%)& 336 (6.72\%) & 1145 (20.99\%)\\\hline
\bf total & \bf 18103 & \bf 4998 & \bf 5588 \\
\hline
\end{tabular}
\end{center}
\caption{Dataset term class amount and relative count.}
\label{termamount} 
\end{table}

\begin{table}[t]
\begin{center}
\begin{tabular}{|c|c|c|c|c|c|c|c|}
\hline
\multirow{2}{*}{nestedness} & \multirow{2}{*}{class} & \multicolumn{3}{|c|}{character length} & \multicolumn{3}{|c|}{word length} \\ \cline{3-8}
& & min & max & mean &  min & max & mean  \\ \hline
\multirow{4}{*}{outermost} & specific & 2 & 91 & 17.58 & 1 & 13 & 1.89 \\ 
& common & 2 & 41 & 10.98 & 1 & 10 & 1.36 \\ 
& nomen & 3 & 95 & 24.06 & 1 & 14 & 3.19 \\ \cline{2-8}
& all & 2 & 95 & 17.58 & 1 & 14 & 1.84 \\ \hline
\multirow{4}{*}{inner} & specific & 1 & 83 & 13.95 & 1 & 13 & 1.46 \\ 
& common & 2 & 31 & 7.54 & 1 & 5 & 1.12 \\ 
& nomen & 3 & 11 & 7.11 & 1 & 3 & 1.22 \\  \cline{2-8}
& all & 1 & 83 & 13.95 & 1 & 13 & 1.32 \\ \hline
\multirow{4}{*}{overall} & specific & 1 & 91 & 17.83 & 1 & 13 & 1.79 \\ 
& common & 2 & 41 & 9.47 & 1 & 10 & 1.25 \\ 
& nomen & 3 & 95 & 23.79 & 1 & 14 & 3.16    \\  \cline{2-8}
& all & 1 & 95 & 15.78 & 1 & 14 & 1.69 \\ \hline
\end{tabular}
\end{center}
\caption{Train dataset terms length.}
\label{termlength} 
\end{table}

In Table \ref{termamount}, the frequencies of each term type in each track are shown. It can be seen that specific terms are the most frequent in the dataset. Table \ref{termlength} demonstrate the length distribution for each term type in the training set. We can see that nomen terms are longest (their length is about 3 words on average), and common terms are shortest (1-2 words on average).

In Tables \ref{nestedness_train}, \ref{nestedness_dev12}, \ref{nestedness_dev3}, \ref{innertoouter}, the levels of term nestedness for different sets and tracks are shown. The maximal level of nestedness in term annotations is 5. The fraction of longest (outmost) entities is about 55-56\%. That is, nested terms constitute a significant share of annotated terms.

\begin{table}[t]
\begin{center}
\begin{tabular}{|c|c|c|c|c|}
\hline
level & specific & common & nomen & total \\ \hline 
1 (outermost) & 9589 (75.72\%) & 2734 (39.07\%) & 564 (96.91\%) & 12887 (55.26\%) \\ \hline
2 & 2690 (21.24\%) & 1734 (24.78\%) & 9 (1.55\%) & 4433 (19.01\%)\\ 
3 & 360 (2.84\%) & 367 (5.24\%)& 0 (0\%)& 727 (3.12\%) \\ 
4 & 25 (0.20\%) & 30 (0.43\%) & 0 (0\%)& 55 (0.24\%)\\
5 & 0 (0\%) & 1 (0.01\%)& 0 (0\%) & 1 (0.01\%) \\ \hline
total (inner) & 3075 (19.54\%) & 2132 (30.47\%) & 9 (1.55\%) & 5216 (22.37\%)\\ \hline
total (overall) & 15739 & 6998 & 582 & 23319 \\ \hline
\end{tabular}
\end{center}
\caption{Nestedness count of train dataset}
\label{nestedness_train} 
\end{table}

\begin{table}[t]
\begin{center}
\begin{tabular}{|c|c|c|c|c|}
\hline
level & specific & common & nomen & total \\ \hline 
1 (outermost) & 2606 (62.52\%) & 749 (41.59\%) & 269 (66.75\%) & 3624 (56.87\%) \\ \hline
2 & 668 (16.03\%) & 423 (23.49\%) & 67 (16.63\%) & 1158 (18.17\%)\\ 
3 & 92 (2.21\%) & 87 (4.83\%)& 0 (0\%)& 179 (2.81\%) \\ 
4 & 18 (0.43\%) & 12 (0.67\%) & 0 (0\%)& 30 (0.47\%)\\
5 & 3  (0.07\%) & 4 (0.22\%)& 0 (0\%) & 7 (0.11\%) \\ \hline
total (inner) & 781 (18.74\%) & 526 (29.21\%) & 67 (16.63\%)  & 1374 (21.56\%)\\ \hline
total (overall) & 4168 & 1801 & 582 & 6372 \\ \hline
\end{tabular}
\end{center}
\caption{Nestedness count of dev dataset (track 1 \& 2)}
\label{nestedness_dev12} 
\end{table}

\begin{table}[t]
\begin{center}
\begin{tabular}{|c|c|c|c|c|}
\hline
level & specific & common & nomen & total \\ \hline 
1 (outermost) & 2404 (58.12\%) & 705 (44.48\%) & 940 (66.86\%) & 4049 (56.81\%) \\ \hline
2 & 722 (17.46\%) & 373 (23.53\%) & 215 (15.29\%) & 1310 (18.38\%)\\ 
3 & 131 (3.17\%) & 57 (3.60\%)& 15 (1.07\%)& 203 (2.85\%) \\ 
4 & 13 (0.31\%) & 10 (0.63\%) & 3 (0.21\%)& 26 (0.36\%)\\
5 & 0 (0\%) & 0 (0\%)& 0 (0\%) & 0 (0\%) \\ \hline
total (inner) & 866 (20.94\%) & 440 (27.76\%) & 233 (16.57\%) & 1539 (21.59\%)\\ \hline
total (overall) & 4136 & 1585 & 1406 & 7127 \\ \hline
\end{tabular}
\end{center}
\caption{Nestedness count of dev dataset (track 3) }
\label{nestedness_dev3} 
\end{table}

\begin{table}[t]
\begin{center}
\begin{tabular}{|c|c|c|c|}
\hline
inner $\backslash$ outer & specific & common & nomen \\ \hline 
specific & 3124 & 3  & 358  \\
common & 1508 & 773 &  281 \\ 
nomen & 4 & 0 & 5  \\ \hline
total  &  4636 & 776  &  644\\ \hline
\end{tabular}
\end{center}
\caption{Inner classes of terms inside outermost terms of train dataset}
\label{innertoouter} 
\end{table}

\section{RuTermEval Competition Solution}

\subsection{Nested Named Entity Extraction}

We depicted the term extraction task of the RuTermEval competition as a named entity extraction task, since both tasks are sequence labeling tasks. Moreover, the term extraction task could be framed as a named entity extraction task on a specific domain, that is, terms in the RuTermEval dataset. Thus, terms are onwards considered entities throughout this paper.

More specifically, in entity recognition tasks, for a given text sequence $X = \{x_1, x_2, \dots, x_n\}$, where $n$ is its length, we need to assign each subsequence $E = \{(x_i, x_{i+1}, \dots, x_j) ~|~ 1 \leqslant i \leqslant j \leqslant n; i, j, n \in \mathbb{N}\}$ to its corresponding label $y_{i,j} \in Y$ (if any), where $Y$ is a predetermined set of entity classes. 

In the usual setting of the ''flat'' entity recognition task $\forall e_1, e_2 \in E: e_1 \nsubseteq e_2$, i.e.,  entities in a sequence cannot contain other entities within them. Moreover, $\forall e_1, e_2 \in E : e_1 \cap e_2 = \varnothing, e_1 \neq e_2$ i.e., such entities do not overlap too. We will denote such labeling as F: $$F = \{e \in E~|~\forall e_1, e_2: e_1 \cap e_2 = \varnothing \}.$$

On the other hand, in practice, entities can overlap. In this case, we are having nested labeling, i.e. $\exists e_1, e_2 \in E : e_1 \subset e_2$. We will denote such data as $N$:
$$N = \{e \in E~|~ \exists ~ e_1, e_2: e_1 \subset e_2; \forall~e_3, e_4:\text{either}~e_3 \subset e_4~\text{or}~e_4 \subset e_3~\text{or}~e_3 \cap e_4 = \varnothing\}.$$
Note: $F \subset N$.

The RuTermEval competition is devoted to the recognition of nested terms. Our solution is based on the Binder model, learned on training data with the ruRoberta-large pretrained language model \cite{zmitrovich2023family}. 

\subsection{Binder Model}

The main model used in the competition and in this study was the \texttt{Binder} model \cite{zhangOptimizingBiEncoderNamed2023}. This model allows for extracting entities (named entities, terms) from sentences based on the so-called descriptions of entities and a contrastive learning method.

The input is the sequence of words $x_1, \dots, x_n$ and the description of the named entity $E_k$ of class $k$. Both are fed to the so-called encoders, which represent the original text data as vector representations (the so-called embeddings), enriched with contextual information about each word. The most popular encoder is a model such as \texttt{BERT} \cite{kenton2019bert}. The \texttt{BERT} model also allows for obtaining vector representation of the whole sentence using a special token \texttt{[CLS]}. 

Each obtained vector is mapped onto a single vector space using a linear layer. In this space, it is assumed that all entities of the same type should be closer to each other than any other entity, as well as any other subsequences of words of the original sequence.

The main idea of the model is  so-called contrastive learning. It consists in the fact that the model learns not from the vector sequences of words and their subsequences but learns to bring entities of one type closer to a certain single center of the most characteristic entity of this type and, conversely, to move away all other uncharacteristic subsequences.

At the prediction stage, the model is required to find out where the boundary between positive and negative mentions is drawn. To do this, based on the \texttt{[CLS]} vector of the second encoder used to encode the main sequence, its position in the same vector space is trained, and the distance between it and the anchor is the desired radius within which the mentions are positive. This is the so-called dynamic threshold approach, since it is trained together with the rest of the model, and is not fixed in advance.

\subsection{Model parameters}

This solution was the same for all three tracks. In the first track, we considered all terms mentions as entities of the dummy class \textit{any}, treating the task of term identification as term classification of a single entity type. In tracks 2 and 3, the Binder model was trained to extract three term types. 

We trained this model on 128 epochs on GPU with batch size of 8, maximum sequence length of 192, doc\_stride of 16, learning rate of $3 * 10^{-5}$ using AdamW optimizer \cite{loshchilov2019decoupledweightdecayregularization} with default settings. These parameters were the same for the solutions for the three tracks. 

In Tables \ref{track1allres}, \ref{track2allres}, \ref{track3allres} we show the results of our submissions, compared to results of other teams.

\begin{table}[t]
\begin{center}
\begin{tabular}{|c|c|c|c|}
\hline
User & Team Name & Scoreboard F1 score \\ \hline 
\bf fulstock (ours) & \bf LAIR RCC MSU & \bf 0.7940 \\
VladSemak & VSemak & 0.7685 \\
ivan\_da\_marya & Ivan da Marya & 0.5619 \\
ragunna & KiPL SPBU & 0.5349 \\
angyling & - & 0.5333 \\
VatolinAlexey & ai & 0.0000 \\
 \hline
\end{tabular}
\end{center}
\caption{All teams results on the Track 1 of RuTermEval competition.}
\label{track1allres} 
\end{table}

\begin{table}[t]
\begin{center}
\begin{tabular}{|c|c|c|c|c|c|}
\hline
User & Team Name & Total score & Weighted F1 score & Class-agnostic F1 score \\ \hline 
\bf fulstock (ours)  & \bf LAIR RCC MSU & \bf 0.6997 & \bf 0.6997 & \bf 0.78 \\
VladSemak & VSemak & 0.6996 & 0.6996 & 0.77 \\
VatolinAlexey & ai & 0.5797 & 0.5797 & 0.63 \\
ragunna & KiPL SPBU & 0.5043 & 0.5043 & 0.52 \\
angyling & - & 0.3137 & 0.3137 & 0.53 \\
 \hline
\end{tabular}
\end{center}
\caption{All teams results on the Track 2 of RuTermEval competition.}
\label{track2allres} 
\end{table}

\begin{table}[t]
\begin{center}
\begin{tabular}{|c|c|c|c|c|}
\hline
User & Team Name & Total score & Weighted F1 score & Class-agnostic F1 score \\ \hline 
\bf fulstock (ours) & \bf LAIR RCC MSU & \bf 0.4823 & \bf 0.4823 & \bf 0.60  \\
VladSemak & VSemak & 0.4654 & 0.4654 & 0.51 \\
angyling & - & 0.4370 & 0.44 & 	0.53 \\
 \hline
\end{tabular}
\end{center}
\caption{All teams results on the Track 3 of RuTermEval competition.}
\label{track3allres} 
\end{table}

We can see that the approach of nested term recognition  based on the Binder model came out first on all three tracks, overtaking other teams submissions. Though, in the second track the next participant got the score almost identical to ours --- with negligibly small difference. Thus, we can consider that our method surpasses the others in first and third tracks, but stays on the same line of prediction performance as the second-best method in the second track. 

\subsection{Error analysis}

We perform error analysis on the predictions of the development dataset. 

First, we can see that the  model confuses common and specific terms, for example: 
\begin{examples}
\selectlanguage{russian}
\item 
    \gll оценке ~ сочетаемости ~ слов
         assessment of compatibility of words
    \glt 
    \glend
\item 
    \gll синтаксического словаря
         syntactic dictionary
    \glt 
    \glend
\selectlanguage{british}
\end{examples}
The model recognized these terms as common, though they were actually labeled as specific. However, the following examples were labeled common, but the model recognized them as specific.
\begin{examples}
\selectlanguage{russian}
\item 
    \gll повтору
         repeat
    \glt 
    \glend
\item 
    \gll модальная
         modal
    \glt 
    \glend
\selectlanguage{british}
\end{examples}
 
The confusion of these classes occurs rather frequently. We believe this is due to their definition: model fails to differ when the found term is used by non-specialists and when it is not. We can also see that in the first track there are no such errors.   

Second, from our results, we see that most of the model prediction errors were false negative, i.e., the model retrieves less amount of erroneous terms than not retrieves the labeled ones. 

Third, we see some labeling ambiguity in the original data. For example, 
\begin{examples}
\selectlanguage{russian}
\item 
    \gll ядро ~ системы ~ персонализированного синтеза ~ речи по тексту
         core of system of personalized synthesis of speech from text
    \glt 
    \glend
\item 
    \gll системы ~ персонализированного синтеза ~ речи
         system of personalized synthesis of speech
    \glt 
    \glend
\item 
    \gll персонализированного синтеза ~ речи
         personalized synthesis of speech
    \glt 
    \glend
\selectlanguage{british}
\end{examples}
were all labeled as specific, but
\begin{examples}
\selectlanguage{russian}
\item 
    \gll системы ~ персонализированного синтеза ~ речи по тексту
         system of personalized synthesis of speech from text
    \glt 
    \glend
\selectlanguage{british}
\end{examples}
was not. The model recognized all four of these examples as specific terms, but because the last was not labeled, it resulted in a false negative error.

\section{Nested Term Recognition from Flat Supervision Task}

Currently, most existing datasets labeled with terms have flat annotations. However, we can see that many terms can be nested. Therefore, we study methods for extracting nested terms from flat-term annotations based on the RuTermEval data.

\subsection{Task}

We consider the problem of extracting nested data given only flat labeling. That is, for training data only flat data $F$ is given, but for validation and test subsets we have full nested data $N$. This task was proposed for named entities in \cite{zhu2022recognizingnestedentitiesflat}.

There are different options available for flat data $F$. For example, if $F$ is generated from nested data $N$ by deleting all overlaps $e_1 \subset e_2, e_1, e_2 \in N$, we can delete only $e_1$ or $e_2$ from $N$ to remove this overlap. Thus, we come to many scenarios of flat data. 

We consider the ultimate and most common option --- ''outermost'' flat data. By ''outermost'' we mean entities that can contain other entities, but are not part of any other entity in data, i.e., ''outermost'' flat data $F = F_u$ is as:
$$ F_u = \{ e \in N~|~ \nexists e': e \subset e' \} $$

In this work, we propose several methods for the prediction of nested data $N$ given only ''outermost'' flat data $F_u$. Furthermore, we denote all remaining nested entities that are not part of outermost flat data as inner entities $I$: $I = N~\backslash~F_u$.

\subsection{Baselines}

Two baselines are used: full nested learning and ''pure'' flat learning.

Full nested learning is the option to learn a model on all available nested data $N$  to achieve the best results in extracting nested entities as described in Section 4. This baseline would give us the upper limit that should be achieved through flat supervision methods.

''Pure'' flat learning is the option of learning only on available flat data $F$ without any changes to the learning scheme, model architecture, or data augmentation. This is the simplest approach in flat supervision and is regarded as a bottom baseline.

Both approaches would give us the upper and lower expected limits.

\subsection{Automatic pseudo-labeling}

In this work, we propose several novel methods grouped by the term ''automatic pseudo-labeling''. All these approaches represent data augmentation techniques that try to leverage the omitted nestedness of the data.

There are four approaches in total, as follows:
\begin{itemize}
    \item ''Inclusions'' (simple and lemmatized); 
    \item ''Damaged cross-validation'' (''early'' and ''late''). 
\end{itemize}

The explanations are given below.

\subsubsection{Inclusions}

Consider the flat dataset $F_u$ defined as above. Entities in such dataset can contain inner entities within them; they were either unlabeled or deleted from initial labeling. Since the given text itself was not altered, the entities are still hidden in the text, an we just cannot know where exactly. 

In practice, many entities do not appear just once in the text. Each occurrence is called ''mention'', which is widely known in other fields such as entity linking. Furthermore, each entity can have many mentions in the text.

This phenomenon inspired the approach of so-called ''inclusions''. We consider the hypothesis that each mention of the known entities should be tackled as entities too. Inclusions are mentions of known entities inside other ones.

We add such inclusions to the initial flat dataset as a pseudo-labeling, much like data augmentation, and train model on a new dataset. Inclusions are computed only at the training phase. 

Another variant of this method additionally exploits the lemmatization of such inclusions. Before we looked for mentions of known entities ''as whole,'' i.e., each mention should be written exactly as the original entity itself. But if we split such entities as a list of words, lemmatize and combine them in one unordered set of lemmas, we can extend the definition of inclusions. 

From our calculations, there were 1296 simple inclusions (and 14183 entities as a result) and 3681 lemmatized inclusions (and 16568 entities as a result) in the train dataset.

\subsubsection{Damaged Cross-prediction}

\begin{figure}[t]
\centering
\begin{subfigure}{.5\textwidth}
  \centering
  \includegraphics[width=\linewidth]{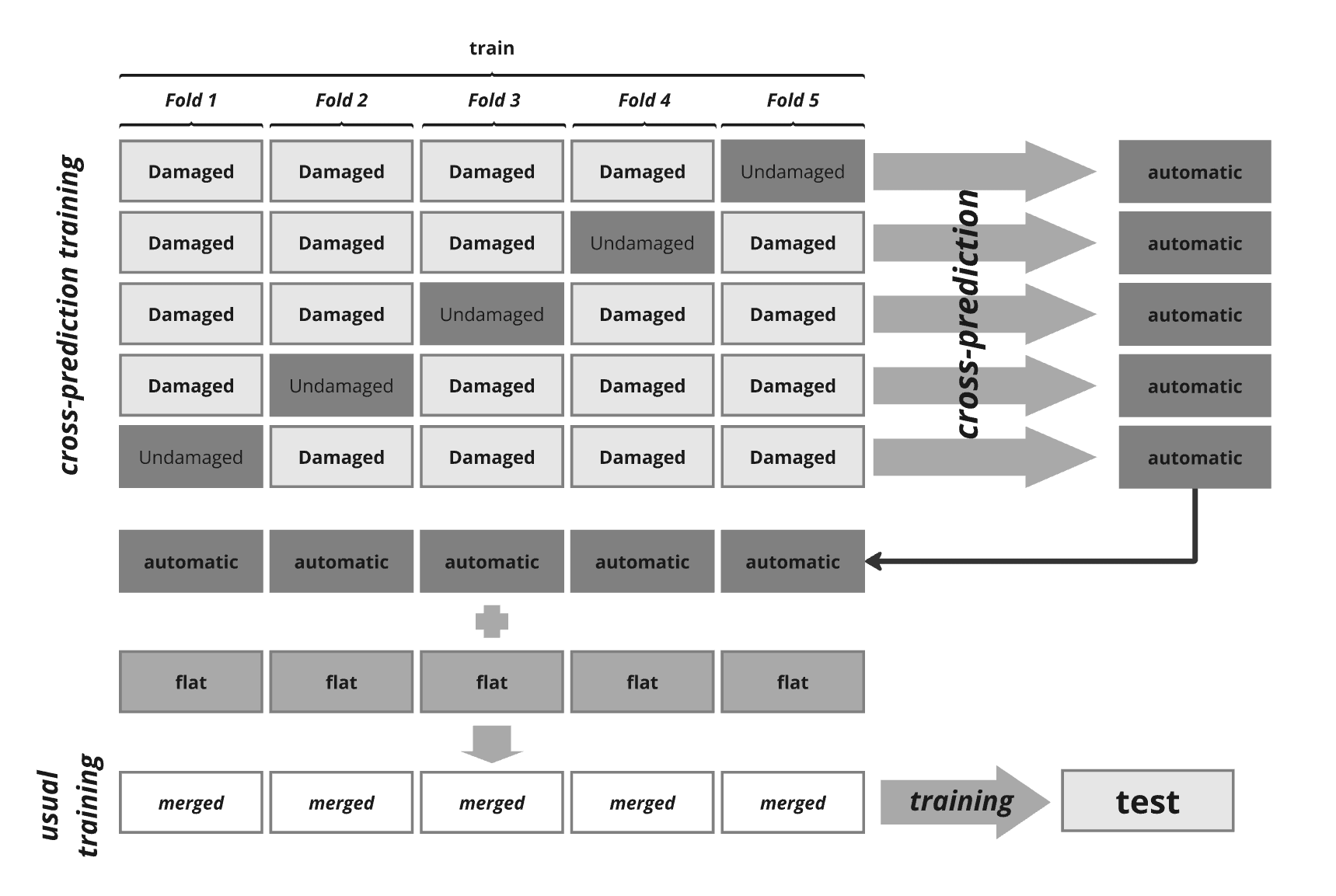}
  \caption{''Early'' damage}
  \label{fig:damage1}
\end{subfigure}%
\begin{subfigure}{.5\textwidth}
  \centering
  \includegraphics[width=\linewidth]{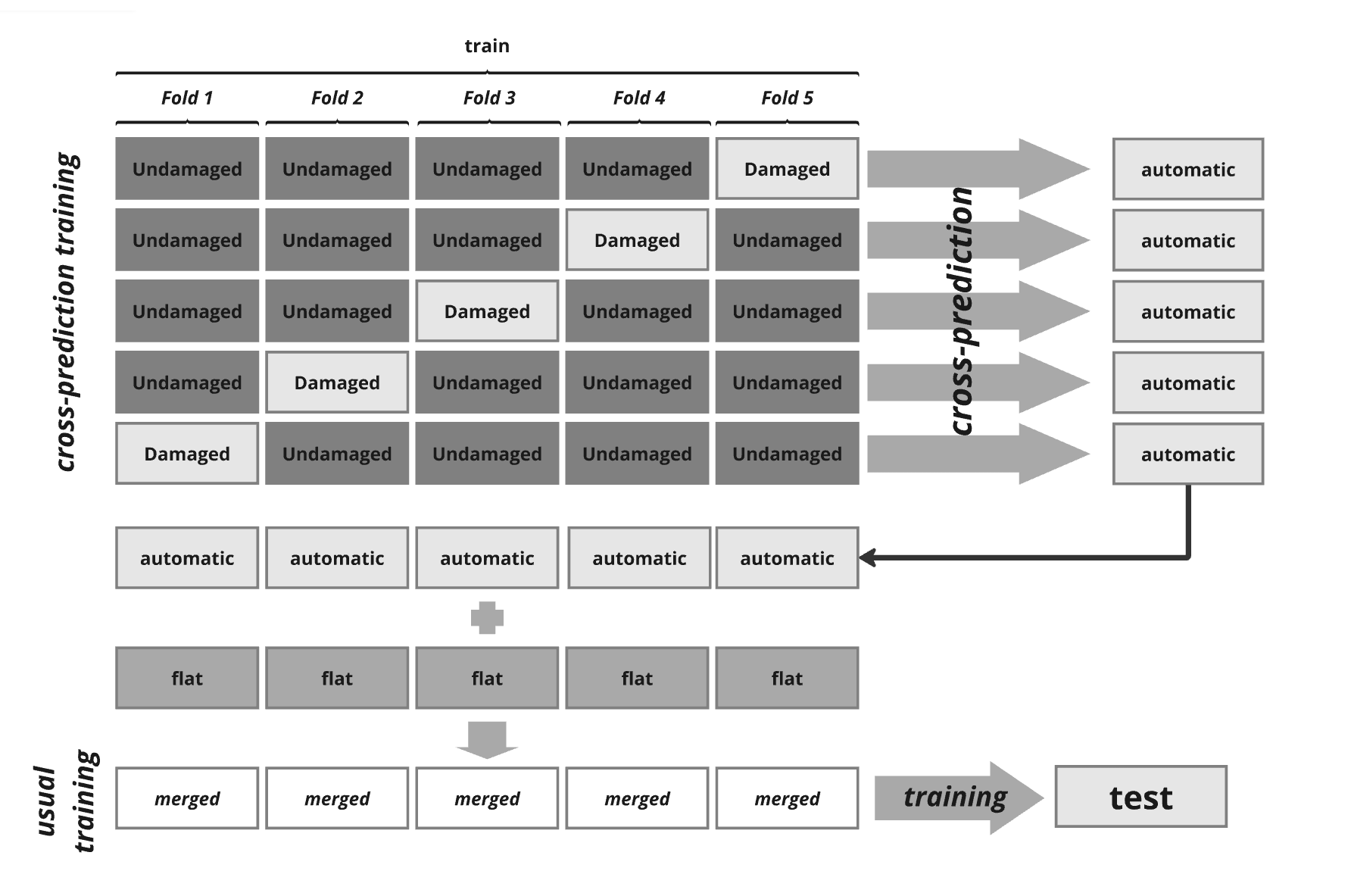}
  \caption{''Late'' damage}
  \label{fig:damage2}
\end{subfigure}
\caption{Scheme of damaged cross-prediction method, example of 5 folds.}
\label{fig:damage}
\end{figure}

Another approach  to pseudo-labeling guides the model during training to ''look for'' more inner entities. 

Firstly, flat training  data is divided into $K$ folds ($K$ is predefined). These folds are used for cross-validation: $K-1$ of them are used as a subtraining dataset, while the remaining fold is used as a subdevelopment dataset. 

Secondly, we change the given flat data or ''damage'' it. Only training subset is changed. Here we propose two different methods:
\begin{itemize}
    \item ''Early'' damage. We delete labeling of all long flat entities (of length 3 and more) from all subtraining folds, while changing one of the words $x_t$ of such entity to some other character sequence. 
    \item ''Late'' damage. Same as ''early'', but we damage the subdevelopment fold, while subtraining folds remain untackled.
\end{itemize}
From our calculations, out of 12887 of train subset entities, 2053 were damaged, while 10834 others were kept the same. 

Thirdly, we train model $K$ times, just like in cross-validation: $K-1$ subtraining folds are used for training, while the remaining fold is used for testing. For this purpose, in both methods the model tries to predict entities in parts of the initial training dataset. After all $K$ training procedures, we get predictions of the model on all $K$ folds. All these predictions are now considered as additional pseudo-labeling. 

Finally, obtained labeling is added to the initial flat training subset. After that, the method stays familiar with usual training and evaluation. Figure \ref{fig:damage} depicts the early damage and late damage pseudo-labeling schemes.

\begin{table}[t]
\begin{center}
\begin{tabular}{|c|c c c|c c c|c|}
\hline & \multicolumn{6}{|c|}{\bf dev} & \bf test \\ \hline
\multirow{2}{*}{\bf Approaches} & \multicolumn{3}{|c|}{\bf F1 micro, \%} & \multicolumn{3}{|c|}{\bf F1 macro, \%} & \multirow{2}{*}{Scoreboard F1 \%} \\ \cline{2-7}
& overall & inner & outer & overall & inner & outer & \\ \hline
pure flat & 64.73 & 0.43 & 71.53 & 66.23 & 0.39 & 73.04 & 65.10 \\ \hline
inclusions & 68.28 & 17.78 & \bf 71.65 & 67.79 & 11.74 & 71.92 & 67.42 \\
lemm. inclusions & 70.51 & 32.69 & 71.28 & 70.47 & 23.50 & 72.24 & 70.26 \\ \hline 
early damage &  69.87 & 23.76 & 71.13 & 72.43 & 25.65 & 72.64 & 71.09 \\ 
late damage & 66.66 & 10.57 & 70.06 & 68.63 & 10.43 & 72.31 & 68.90 \\ \hline
lemm. inc. + early dmg & \bf 71.99 & \bf 38.42 & 70.52 & \bf 74.30 & \bf 36.04 & 73.18 & \bf 72.81 \\
lemm. inc. + late dmg &  71.16 & 32.88 & 71.56 & 71.91 & 28.25 & \bf 73.26 & 71.89 \\ \hline
full &  79.87 & 67.73 & 72.90 & 81.84 & 65.20 & 74.68 & 79.40 \\
\hline
\end{tabular}
\end{center}
\caption{Results on track 1.}
\label{track1res} 
\end{table}

\begin{table}[t]
\begin{center}
\begin{tabular}{|c|c c c|c c c|c c|}
\hline & \multicolumn{6}{|c|}{\bf dev} & \multicolumn{2}{c|}{\bf test} \\ \hline
\multirow{2}{*}{\bf Approaches} & \multicolumn{3}{|c|}{\bf F1 micro, \%} & \multicolumn{3}{|c|}{\bf F1 macro, \%} & \multirow{2}{*}{w. F1, \%} & \multirow{2}{*}{c/a F1, \%} \\ \cline{2-7}
& overall & inner & outer & overall & inner & outer & & \\ \hline
pure flat & 61.01 & 0.68 & 67.04 & 63.61 & 1.42 & 69.62 & 55.04 & 63.60 \\ \hline
inclusions & 64.98 & 17.79 & \bf 67.96 & 65.50 & 12.97 & 69.28 & 58.64 & 66.60 \\
lemm. inclusions & 66.84 & 30.80 & 67.55 & 67.02 & 24.06 & 68.60 & 59.49 & 70.21 \\ \hline 
early damage & 65.34 & 19.34 & 65.32 & 66.96 & 20.86 & 66.97 & 60.39 & 70.03 \\ 
late damage & 63.18 & 8.63 & 66.68 & 64.51 & 8.68 & 68.17 & 58.84 & 67.29 \\ \hline
lemm. inc. + early dmg  & \bf 68.64 & \bf 36.40 & 66.40 & \bf 69.52 & \bf 32.15 & 68.20 & \bf 63.10 & \bf 73.37 \\
lemm. inc. + late dmg  &  67.13 & 32.97 & 66.66 & 68.31 & 28.07 & \bf 69.74 & 60.97 & 71.41 \\ \hline
full & 76.06 & 64.76 & 68.44 & 77.40 & 61.29 & 70.41 & 69.97 & 77.79 \\
\hline
\end{tabular}
\end{center}
\caption{Results on track 2.}
\label{track2res} 
\end{table}

\begin{table}[t]
\begin{center}
\begin{tabular}{|c|c c c|c c c|c c|}
\hline & \multicolumn{6}{|c|}{\bf dev} & \multicolumn{2}{c|}{\bf test} \\ \hline
\multirow{2}{*}{\bf Approaches} & \multicolumn{3}{|c|}{\bf F1 micro, \%} & \multicolumn{3}{|c|}{\bf F1 macro, \%} & \multirow{2}{*}{w. F1, \%} & \multirow{2}{*}{c/a F1, \%} \\ \cline{2-7}
& overall & inner & outer & overall & inner & outer & & \\ \hline
pure flat & 40.20 & 1.79 & 42.36 & 36.76 & 1.66 & 38.67 & 40.17 & 51.58 \\ \hline
inclusions & 43.99 & 9.99 & 44.36 & 39.78 & 7.78 & 40.61 & 42.05 & 55.54 \\
lemm. inclusions & 47.17 & 18.32 & 45.10 & 41.65 & 14.42 & 40.80 & 43.69 & 57.23 \\ \hline 
early damage &  45.84 & 13.02 & 44.44 & 40.85 & 9.98 & 40.68 & 43.39 & 55.99\\ 
late damage &  37.21 & 2.08 & 39.26 & 33.88 & 1.96 & 35.95 & 41.02 & 54.10 \\ \hline
lemm. inc. + early dmg  &  \bf 48.62 & \bf 20.49 & \bf 46.19 & \bf 43.52 & \bf 15.86 & \bf 42.50 & \bf 45.47 & \bf 58.75\\
lemm. inc. + late dmg  &  46.77 & 17.50 & 45.07 & 41.16 & 13.79 & 40.50 & 43.95 & 56.94 \\ \hline
full & 50.35 & 29.57 & 46.18 & 45.16 & 23.91 & 42.18 & 48.23 & 60.38 \\
\hline
\end{tabular}
\end{center}
\caption{Results on track 3.}
\label{track3res} 
\end{table}

All approaches were conducted in all three tracks, both on development and test subsets. Results are given in \ref{track1res}, \ref{track2res}, \ref{track3res}. In the RuTermEval competition in the second and third tracks two metrics were utilized namely weighted F1 (denoted as w. F1) and class-agnostic F1 (denoted as c/a F1).

First, we can see in all three tracks that pure flat approach extracts almost none of inner entities, while still capable of extracting outermost entities, though underperforming. 

Second, as first two tracks consider same data but with different labeling, we see similar results on all approaches. But when task switches from term identification to term classification, all methods perform a bit poorer in general. Moreover, the third track considers the prediction on the different data domain --- hence worse results in comparison. 

Third, we can see that approaches on development and test sets share similar relative results.

Fourth, we see that all approaches enhance the performance of the inner terms prediction, hence the better result in the competition too. 

Fifth, inclusions achieve good boost performance with its pseudo-labeling, from $0.68\%$ to $17.79\%$ inner terms prediction on second track. Moreover, lemmatized inclusions boost the results even more --- up to $30.80\%$. 

Sixth, damage cross-prediction methods achieve results around the same level of inclusions. We see that the ''late'' damage appeared to be poorer in general: model could not really retrieve damaged entities. We believe that this is due change of the context and hence embeddings confusion. 

Seventh, merging together results of damaged cross-prediction and lemmatized inclusions, we achieved best inner terms predictions performance, up to $36.40\%$. On third track, this approach almost achieves the full nested approach, $58.75\%$ to $60.38\%$. Thus, we see that such simple pseudo-labeling can be both very helpful and viable for recognizing nested terms from flat data. 

\subsection{Error analysis}

As in our RuTermEval solution, we perform the same error analysis of our methods on predictions on the development dataset. 

Firstly, we see that the model receives the ambiguity of the predictions because of nestedness. For example,
\begin{examples}
\selectlanguage{russian}
\item 
    \gll лексической системы ~ языка
         lexical system of language
    \glt is labeled as specific and model did not recognize it at all, but
    \glend
\item 
    \gll лексической системы
         lexical system
    \glt was recognized by model as specific term, though it was not labeled in data.
    \glend
\selectlanguage{british}
\end{examples}

Secondly, we see another common pattern. Due to much less nested data in all our approaches with pseudo-labeling, model still fails to generalize on inner terms. Nevertheless, with our approach model now does recognize some flat terms that the usual nested approach of our best solution failed to extract instead:
\begin{examples}
\selectlanguage{russian}
\item 
    \gll озимой пшеницы
         winter wheat
    \glt was extracted correctly in flat approaches, while nested solution did not recognize it (though it was labeled);
    \glend
\item 
    \gll пшеницы
         wheat
    \glt 
    \glend
\item 
    \gll зерна
         grain
    \glt were mistakenly extracted as terms by flat approaches, though they were actually not (i.e., not labeled so).
    \glend
\selectlanguage{british}
\end{examples}

We believe that the model got better trained on longer entities, which resulted in such prediction behaviour. 

Still, this hypothesis holds true to erroneous predictions: for longer subsequences where there was no term, the model retrieved them in flat approaches, while the best nested solution did not:
\begin{examples}
\selectlanguage{russian}
\item 
    \gll биологического биоцидного препарата для ~ борьбы с вредными членистоногими
         biological biocidal preparation for the control of harmful arthropods
    \glt 
    \glend
\selectlanguage{british}
\end{examples}
was mistakenly extracted as a term in flat approaches. We see that such errors were the most frequent.   

Thirdly, our pseudo-labeling inclusions approach introduced to the trained model that many mentions of some term are always terms too. Thus, there are many mistakes such as:
\begin{examples}
\selectlanguage{russian}
\item 
    \gll почва
         soil
    \glt at positions 239, 243; 901, 906; 1265, 1270; 1304, 1309, etc. in text track3-test1-81
    \glend
\item 
    \gll урожайность
         productivity of land
    \glt at positions 132, 143; 298, 309; 437, 448; 567, 578; 641, 652, etc. in text track3-test1-81
    \glend
\selectlanguage{british}
\end{examples}
while nested solution did not produce them. 

\section{Conclusion}

In this work, we present our solution to the RuTermEval competition to all three tracks. Our solution via regarding Nested Term Extraction task as a Nested Named Entity Extraction task appeared to be the most effectively or at least alongside other competitors' solutions --- we have got the first place in all three tracks, devoted to different scenarios of term extraction. We used the Binder model, which was previously successfully applied to the recognition of nested named entities, to extract nested terms and obtained the best results of term recognition in all three tracks of the RuTermEval competition.

We describe and motivate a new task --- Nested
Term Recognition from Flat Supervision. In this task, a model should predict nested terms based only on flat labeling at hand. We propose some approaches to this task and evaluate them in the RuTermEval competition. Of them, the combined lemmatized inclusions with early damage approaches resulted in the best inner term prediction score, coming close to the score on the full nested data. 

\bibliography{dialogue.bib}

\begin{thebibliography}{}

\bibitem[\protect\citename{Astrakhantsev \bgroup et al.\egroup }2015]{astrakhantsev2015methods}
Nikita~A Astrakhantsev, Denis~G Fedorenko, and D~Yu Turdakov.
\newblock 2015.
\newblock Methods for automatic term recognition in domain-specific text collections: A survey.
\newblock {\em Programming and Computer Software}, 41:336--349.

\bibitem[\protect\citename{Bolshakova and Efremova}2015]{bolshakova2015heuristic}
Elena~I Bolshakova and Natalia~E Efremova.
\newblock 2015.
\newblock A heuristic strategy for extracting terms from scientific texts.
\newblock  // {\em Analysis of Images, Social Networks and Texts: 4th International Conference, AIST 2015, Yekaterinburg, Russia, April 9--11, 2015, Revised Selected Papers 4}, P 297--307. Springer.

\bibitem[\protect\citename{Bolshakova and Semak}2024]{bolshakovaexperimental}
Elena~I Bolshakova and Vladislav~V Semak.
\newblock 2024.
\newblock An experimental study on cross-domain transformer-based term recognition for russian.
\newblock  // {\em Proceedings of DAMDID-2024}.

\bibitem[\protect\citename{Bolshakova \bgroup et al.\egroup }2013]{bolshakova2013topic}
Elena Bolshakova, Natalia Loukachevitch, and Michael Nokel.
\newblock 2013.
\newblock Topic models can improve domain term extraction.
\newblock  // {\em Advances in Information Retrieval: 35th European Conference on IR Research, ECIR 2013, Moscow, Russia, March 24-27, 2013. Proceedings 35}, P 684--687. Springer.

\bibitem[\protect\citename{Bouma}2009]{bouma2009normalized}
Gerlof Bouma.
\newblock 2009.
\newblock Normalized (pointwise) mutual information in collocation extraction.
\newblock {\em Proceedings of GSCL}, 30:31--40.

\bibitem[\protect\citename{Charalampakis \bgroup et al.\egroup }2016]{charalampakisComparisonSemisupervisedSupervised2016}
Basilis Charalampakis, Dimitris Spathis, Elias Kouslis, and Katia Kermanidis.
\newblock 2016.
\newblock A comparison between semi-supervised and supervised text mining techniques on detecting irony in greek political tweets.
\newblock {\em Engineering Applications of Artificial Intelligence}, 51:50--57, May.

\bibitem[\protect\citename{Hazem \bgroup et al.\egroup }2022]{hazem2022cross}
Amir Hazem, M{\'e}rieme Bouhandi, Florian Boudin, and B{\'e}atrice Daille.
\newblock 2022.
\newblock Cross-lingual and cross-domain transfer learning for automatic term extraction from low resource data.
\newblock  // {\em Proceedings of the Thirteenth Language Resources and Evaluation Conference}, P 648--662.

\bibitem[\protect\citename{Kenton and Toutanova}2019]{kenton2019bert}
Jacob Devlin Ming-Wei~Chang Kenton and Lee~Kristina Toutanova.
\newblock 2019.
\newblock Bert: Pre-training of deep bidirectional transformers for language understanding.
\newblock  // {\em Proceedings of naacL-HLT}, volume~1. Minneapolis, Minnesota.

\bibitem[\protect\citename{Lang \bgroup et al.\egroup }2021]{lang2021transforming}
Christian Lang, Lennart Wachowiak, Barbara Heinisch, and Dagmar Gromann.
\newblock 2021.
\newblock Transforming term extraction: Transformer-based approaches to multilingual term extraction across domains.
\newblock  // {\em Findings of the Association for Computational Linguistics: ACL-IJCNLP 2021}, P 3607--3620.

\bibitem[\protect\citename{Liu \bgroup et al.\egroup }2020]{liuFinBERTPretrainedFinancial2020}
Zhuang Liu, Degen Huang, Kaiyu Huang, Zhuang Li, and Jun Zhao.
\newblock 2020.
\newblock {{FinBERT}}: {{A Pre-trained Financial Language Representation Model}} for {{Financial Text Mining}}.
\newblock  // {\em Proceedings of the {{Twenty-Ninth International Joint Conference}} on {{Artificial Intelligence}}}, P 4513--4519, Yokohama, Japan, July. International Joint Conferences on Artificial Intelligence Organization.

\bibitem[\protect\citename{Loshchilov and Hutter}2019]{loshchilov2019decoupledweightdecayregularization}
Ilya Loshchilov and Frank Hutter.
\newblock 2019.
\newblock Decoupled weight decay regularization.

\bibitem[\protect\citename{Loukachevitch and Nokel}2013]{loukachevitch2013experimental}
Natalia Loukachevitch and Michael Nokel.
\newblock 2013.
\newblock An experimental study of term extraction for real information-retrieval thesauri.
\newblock  // {\em Proceedings of TIA}, P 69--76.

\bibitem[\protect\citename{Loukachevitch \bgroup et al.\egroup }2024]{loukachevitch2024nerel}
Natalia Loukachevitch, Ekaterina Artemova, Tatiana Batura, Pavel Braslavski, Vladimir Ivanov, Suresh Manandhar, Alexander Pugachev, Igor Rozhkov, Artem Shelmanov, Elena Tutubalina, et~al.
\newblock 2024.
\newblock Nerel: a russian information extraction dataset with rich annotation for nested entities, relations, and wikidata entity links.
\newblock {\em Language Resources and Evaluation}, 58(2):547--583.

\bibitem[\protect\citename{Loukachevitch}2012]{loukachevitch2012automatic}
Natalia~V Loukachevitch.
\newblock 2012.
\newblock Automatic term recognition needs multiple evidence.
\newblock  // {\em LREC}, P 2401--2407.

\bibitem[\protect\citename{Marciniak and Mykowiecka}2015]{marciniak2015nested}
Malgorzata Marciniak and Agnieszka Mykowiecka.
\newblock 2015.
\newblock Nested term recognition driven by word connection strength.
\newblock {\em Terminology. International Journal of Theoretical and Applied Issues in Specialized Communication}, 21(2):180--204.

\bibitem[\protect\citename{Nadif and Role}2021]{nadifUnsupervisedSelfsupervisedDeep2021}
Mohamed Nadif and Fran{\c c}ois Role.
\newblock 2021.
\newblock Unsupervised and self-supervised deep learning approaches for biomedical text mining.
\newblock {\em Briefings in Bioinformatics}, 22(2):1592--1603, March.

\bibitem[\protect\citename{Rigouts~Terryn \bgroup et al.\egroup }2022]{rigouts2022acter}
Ayla Rigouts~Terryn, Veronique Hoste, and Els Lefever.
\newblock 2022.
\newblock Acter 1.5: Annotated corpora for term extraction research.
\newblock  // {\em CLARIN Annual Conference Proceedings, 2022}, P 1--4.

\bibitem[\protect\citename{{\v{S}}ajatovi{\'c} \bgroup et al.\egroup }2019]{vsajatovic2019evaluating}
Antonio {\v{S}}ajatovi{\'c}, Maja Buljan, Jan {\v{S}}najder, and Bojana~Dalbelo Ba{\v{s}}i{\'c}.
\newblock 2019.
\newblock Evaluating automatic term extraction methods on individual documents.
\newblock  // {\em Proceedings of the Joint Workshop on Multiword Expressions and WordNet (MWE-WN 2019)}, P 149--154.

\bibitem[\protect\citename{Tran \bgroup et al.\egroup }2024]{tran2024can}
Hanh Thi~Hong Tran, Matej Martinc, Andraz Repar, Nikola Ljube{\v{s}}i{\'c}, Antoine Doucet, and Senja Pollak.
\newblock 2024.
\newblock Can cross-domain term extraction benefit from cross-lingual transfer and nested term labeling?
\newblock {\em Machine Learning}, 113(7):4285--4314.

\bibitem[\protect\citename{Vo \bgroup et al.\egroup }2022]{vo2022automatic}
Chau Vo, Tru Cao, Ngoc Truong, Trung Ngo, and Dai Bui.
\newblock 2022.
\newblock Automatic medical term extraction from vietnamese clinical texts.
\newblock {\em Terminology}, 28(2):299--327.

\bibitem[\protect\citename{Xie \bgroup et al.\egroup }2022]{xiePretrainedLanguageModels2022}
Qianqian Xie, Jennifer~Amy Bishop, Prayag Tiwari, and Sophia Ananiadou.
\newblock 2022.
\newblock Pre-trained language models with domain knowledge for biomedical extractive summarization.
\newblock {\em Knowledge-Based Systems}, 252:109460, September.

\bibitem[\protect\citename{Zhang \bgroup et al.\egroup }2023]{zhangOptimizingBiEncoderNamed2023}
Sheng Zhang, Hao Cheng, Jianfeng Gao, and Hoifung Poon.
\newblock 2023.
\newblock Optimizing {{Bi-Encoder}} for {{Named Entity Recognition}} via {{Contrastive Learning}}, February.

\bibitem[\protect\citename{Zhu \bgroup et al.\egroup }2022a]{zhuRecognizingNestedEntities2022}
Enwei Zhu, Yiyang Liu, Ming Jin, and Jinpeng Li.
\newblock 2022a.
\newblock Recognizing {{Nested Entities}} from {{Flat Supervision}}: {{A New NER Subtask}}, {{Feasibility}} and {{Challenges}}, November.

\bibitem[\protect\citename{Zhu \bgroup et al.\egroup }2022b]{zhu2022recognizingnestedentitiesflat}
Enwei Zhu, Yiyang Liu, Ming Jin, and Jinpeng Li.
\newblock 2022b.
\newblock Recognizing nested entities from flat supervision: A new ner subtask, feasibility and challenges.

\bibitem[\protect\citename{Zmitrovich \bgroup et al.\egroup }2023]{zmitrovich2023family}
Dmitry Zmitrovich, Alexander Abramov, Andrey Kalmykov, Maria Tikhonova, Ekaterina Taktasheva, Danil Astafurov, Mark Baushenko, Artem Snegirev, Tatiana Shavrina, Sergey Markov, Vladislav Mikhailov, and Alena Fenogenova.
\newblock 2023.
\newblock A family of pretrained transformer language models for russian.

\end{thebibliography}
\bibliographystyle{dialogue}

\end{document}